%% file: aaai23.tex
\documentclass[letterpaper]{article} 
\usepackage[submission]{aaai23}  
\usepackage{times}  
\usepackage{helvet}  
\usepackage{courier}  
\usepackage[hyphens]{url}  
\usepackage{graphicx} 
\usepackage{natbib}  
\usepackage{caption} 
\usepackage{algorithm}
\usepackage{algorithmic}
\usepackage{todonotes}

\newcommand{\viz}{\emph{viz.}}

\newcommand{\ie}{{\em i.e.}}

\usepackage{newfloat}
\usepackage{listings}
\DeclareCaptionStyle{ruled}{labelfont=normalfont,labelsep=colon,strut=off} 
\floatstyle{ruled}
\newfloat{listing}{tb}{lst}{}
\floatname{listing}{Listing}
\input{macros}

\title{Speeding up NAS with Adaptive Subset Selection}
\author{
    Vishak Prasad C\textsuperscript{\rm 1},
    Colin White\textsuperscript{\rm 2},
    Paarth Jain\textsuperscript{\rm 1},
    Sibasis Nayak\textsuperscript{\rm 1},
    Ganesh Ramakrishnan\textsuperscript{\rm 1}
}
\affiliations{
    \textsuperscript{\rm 1}\{vishak, paarthjain, sibasisnayak, ganesh\}@cse.iitb.ac.in\\
    \textsuperscript{\rm 2}colin@abacus.ai

}

\usepackage{bibentry}

\begin{document}

\maketitle

\input{abstract}
\input{introduction}

\input{relatedwork}

\input{methodology}
\input{experiments}

\input{conclusion}

\bibliography{04_aaai_2023/aaai23}
\appendix

\input{appendix}

\end{document}

%% file: macros.tex
\usepackage{adjustbox}
\usepackage{comment}
\usepackage{amsmath,amsthm,amssymb}
\usepackage{algorithm}
\usepackage{algorithmic}
\usepackage{multicol}
\usepackage{mathtools}
\usepackage{caption}
\usepackage{float}
\usepackage{graphicx}
\usepackage{graphicx}
\usepackage{subcaption}
\usepackage{wrapfig}
\usepackage{multirow}
\usepackage{mathrsfs}
\usepackage{newfloat}
\usepackage{relsize}
\usepackage{enumitem}
\usepackage{nicefrac}
\usepackage{pifont}
\usepackage{bm}
\usepackage{tikz}
\usetikzlibrary{shapes.misc,shadows}
\usepackage{subcaption}

\newcommand{\dpt}{\mbox{\textsc{Darts-pt}}} 
\newcommand{\dptent}{\mbox{\textsc{Darts-pt-entropy}}} 
\newcommand{\dptrand}{\mbox{\textsc{Darts-pt-rand}}} 
\newcommand{\dptfl}{\mbox{\textsc{Darts-pt-fl}}} 
\newcommand{\model}{\mbox{\textsc{Adaptive-Dpt}}} 
\newcommand{\adehb}{\mbox{\textsc{Adaptive-DEHB}}} 
\newcommand{\abohb}{\mbox{\textsc{Adaptive-BOHB}}}

\DeclareMathOperator*{\argmin}{arg\,min}

\newcommand{\U}{\mathcal{U}}
\newcommand{\V}{\mathcal{V}}
\newcommand{\Lval}{\mathcal{L}_{\text{val}}}
\newcommand{\Ltrain}{\mathcal{L}_{\text{train}}}

%% file: abstract.tex
\begin{abstract}
A majority of recent developments in neural architecture search (NAS) have been aimed at decreasing the computational cost of various techniques without affecting their final performance. Towards this goal, several low-fidelity and performance prediction methods have been considered, including those that train only on subsets of the training data. In this work, we present an adaptive subset selection approach to NAS and present it as complementary to state-of-the-art NAS approaches. We uncover a natural connection between one-shot NAS algorithms and adaptive subset selection and devise an algorithm that makes use of state-of-the-art techniques from both areas. We use these techniques to substantially reduce the runtime of DARTS-PT (a leading one-shot NAS algorithm), as well as BOHB and DEHB (leading multi-fidelity optimization algorithms), without sacrificing accuracy. Our results are consistent across multiple datasets, and towards full reproducibility, we release our code at \url{https://anonymous.4open.science/r/SubsetSelection_NAS-B132}.
\end{abstract}

%% file: introduction.tex
\section{Introduction} \label{sec:introduction}
Neural architecture search (NAS), the process of automating the design of high-performing neural architectures, has been used to discover architectures that outpace the best human-designed 
neural networks~\citep{dai2020fbnetv3, efficientnets, real2019regularized, nas-survey}.
Early NAS algorithms used black-box optimization methods such as reinforcement learning~\citep{zoph2017neural, pham2018efficient} and Bayesian optimization~\citep{nasbot}. A majority
of recent developments has focused on decreasing the cost of NAS without sacrificing performance.

Toward this direction, `one-shot' methods improve the search efficiency by training just a single over-parameterized neural network (supernetwork)~\citep{darts, bender2018understanding}. For example, the popular DARTS~\citep{darts} algorithm applies a continuous relaxation to the architecture parameters, allowing the architecture parameters and the weights to be simultaneously optimized via gradient descent.
While many follow-up works have improved the performance of DARTS~\citep{wang2021rethinking,prdarts, zela2020understanding},
the algorithms are still slow and require computational resources that are expensive in terms of budget and 
CO2 emissions~\citep{tornede2021towards}.

On the other hand, the field of subset selection for efficient machine learning-based model training has seen a flurry of activity. In this area of study, facility location~\citep{mirzasoleiman2020coresets}, clustering~\citep{clark2020electra}, and other
subset selection algorithms are used to select a representative subset of the training data, substantially reducing the runtime of model training.
Recently, adaptive subset selection algorithms have been used to speed up model training  even further~\citep{killamsetty2020glister, killamsetty2021grad}.
Adaptive subset selection is a powerful technique which regularly updates the current subset of the data as the search progresses, to ensure that the performance of the model is maintained.

\begin{figure*}
    \includegraphics[width=0.95\textwidth]{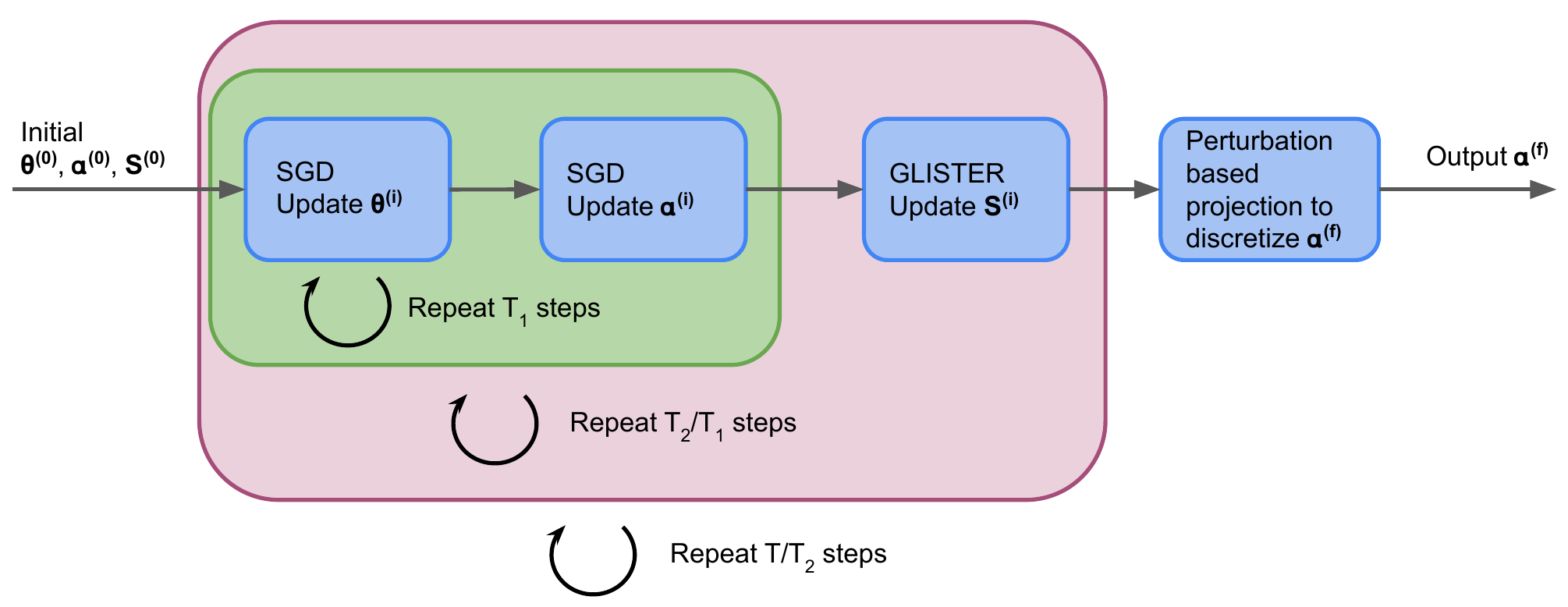}
    \caption{Overview of \model. The algorithm starts with the initial set of weights $\theta^{(0)}$, architecture-parameters $\alpha^{(0)}$, and subset of the training data $S^{(0)}$.
    Throughout the search, the weights $\theta^{(i)}$ and architecture-parameters $\alpha^{(i)}$ are updated with SGD, and the subset $S^{(i)}$ is updated with GLISTER, according to different time schedules. Then the final architecture $\alpha^{(f)}$ is discretized and returned.} 
    \label{fig:overview}
\end{figure*}

In this work, we combine state-of-the-art techniques from both adaptive subset selection and NAS to devise new algorithms.
First, we uncover a natural connection between one-shot NAS algorithms and adaptive subset selection: DARTS-PT \citep{wang2021rethinking} (a leading one-shot algorithm) and GLISTER \citep{killamsetty2020glister} (a leading adaptive subset selection algorithm) are both
cast as bi-level optimization problems on the training and validation sets, 
allowing us to formulate a combined approach, \viz, \model, 
as a mixed discrete and continuous bi-level optimization problem (see Figure \ref{fig:overview} for an overview).
Next, we also combine GLISTER with BOHB \citep{falkner2018bohb} and DEHB \citep{awad2021dehb}, two leading multi-fidelity optimization approaches, to devise \abohb{} and \adehb{}, respectively. 
Across several search spaces, we show that the resulting algorithms achieve significantly improved runtime, without sacrificing performance.
Specifically, due to the use of adaptive subset selection, the training data can be reduced to 10\% of the full training set size, resulting in an order of magnitude decrease in runtime, without sacrificing accuracy. 
To validate these approaches, we compare against baselines such as facility location, entropy-based subset selection \citep{na2021accelerating}, and random subset selection.
Facility location itself is a novel baseline for NAS applications; the codebase we release, that includes four different subset selection algorithms integrated into one-shot NAS, may be of independent interest.

\noindent\textbf{Our contributions.}
We summarize our main contributions.
\begin{itemize}
\item We introduce \model{}, the first NAS algorithm to make use of adaptive subset selection.
The training time needed to find high-performing architectures is substantially reduced.
We also add facility location as a novel baseline for 
subset selection applied to NAS. ({\em c.f.} Section~\ref{sec:methodology}). We extend our idea to show adaptive subset selection complements hyperparameter optimization algorithms using \abohb{} and \adehb{}.
\item Through extensive experiments, we show that \model{}, \abohb{}, and \adehb{} substantially reduces the runtime
needed for running DARTS-PT, BOHB, and DEHB, respectively, with no decrease in the final (test) accuracy of the returned architecture  ({\em c.f.} Section~\ref{sec:experiments}). For reproducibility, we release all of our code.

\end{itemize}

%% file: relatedwork.tex
\section{Related Work} \label{sec:relatedwork}

\paragraph{Neural architecture search.}
NAS has been studied since the 1980s~\citep{dress1987darwinian, tenorio1988self, miller1989designing, kitano1990designing, angeline1994evolutionary}
and has been revitalized in the last few years~\citep{zoph2017neural, darts}.
The initial set of approaches focused on evolutionary search~\citep{stanley2002evolving, real2019regularized}, reinforcement learning~\citep{zoph2017neural, pham2018efficient},
and Bayesian optimization~\citep{nasbot, bananas}.
More recent trends have focused on reducing the computational complexity of
NAS using various approaches.
One line of work aims to predict the performance of neural architectures before they are fully trained, through low-fidelity estimates such as training for fewer epochs~\citep{zhou2020econas, ru2020revisiting},
learning curve extrapolation~\citep{domhan2015speeding, nasbenchx11},
or `zero-cost proxies'~\citep{mellor2020neural, abdelfattah2021zerocost}.


Another line of work takes a \emph{one-shot} approach by representing the entire space of neural architectures by a single `supernetwork', and then performing gradient descent to efficiently 
converge to a high-performing architecture~\citep{darts}.
Since the release of the original differentiable architecture search method~\citep{darts},
several follow-up works have attempted to improve its performance~\citep{darts+, pcdarts, prdarts, li2020geometry, zela2020understanding}.
Recently,  \citet{wang2021rethinking} introduced a more reliable perturbation-based operation scoring technique while computationally returning the final architecture, yielding more accurate models compared to DARTS.

\paragraph{Subset selection.}
Several approaches have been developed in the field of
subset selection for efficient 
model training.
Popular fixed subset selection methods include coreset algorithms~\citep{har2004coresets, mirzasoleiman2020coresets}, facility location~\citep{mirzasoleiman2020coresets}, and entropy-based methods~\citep{na2021accelerating}.
Recently, \citet{killamsetty2020glister} proposed GLISTER
 as an adaptive subset selection method based on a greedy search; an adaptive gradient-matching algorithm for subset selection was also subsequently proposed~\citep{killamsetty2021grad}.

\paragraph{Subset selection in NAS.}
A few existing works have applied (offline) subset selection to the field of NAS.
\citet{na2021accelerating} consider three
subset selection approaches: forgetting events, $k$-center, and entropy-based techniques, showing that the entropy-based approaches result in the best speedup in the case of DARTS.
\citet{shim2021core} consider core-set sampling to speed up PC-DARTS by a factor of 8.
Some more recent 
work~\citep{killamsetty2022automata} employs subset selection
algorithms to obtain greater speed-ups over multi-fidelity methods such as Hyperband~\citep{hyperband} and ASHA~\citep{li2018massively}.
Finally, another league of recent work uses a generative model to create a small set of~\emph{synthetic} training data, which in turn is used to efficiently train architectures during NAS~\citep{such2019generative, rawal2020synthetic}.

%% file: methodology.tex
\section{Methodology} \label{sec:methodology}

\paragraph{Preliminaries.}
We begin by reviewing the ideas behind DARTS and DARTS-PT.
The DARTS search space consists of cells, with each  cell expressed as a directed acyclic graph,
where each edge $(i,j)$ can take on choices of operations $o^{(i,j)}$ such as \texttt{max\_pool\_3x3} or \texttt{sep\_conv\_5x5}.
Let us denote the entire set of possible operations by $\mathcal{O}$.
Each choice of operation for a given edge $(i,j)$, has a corresponding continuous variable $\alpha^{(i,j)}$.
Let $\U$ and $\V$ denote the training and validation sets respectively.
Further, let us denote  the training and validation losses by $\Ltrain$ and $\Lval$ respectively.
For any given dataset, these losses are a function of the architecture parameters and the architecture itself.


DARTS and DARTS-PT are both gradient-based optimization methods that train a supernetwork consisting of weights $\theta$ and architecture-parameters $\alpha$. We will hereafter refer to $\alpha$s as NAS-parameters. 
Each edge in the DARTS search space is given all possible choices ($\mathcal{O}$) for operations, resulting in a mixed output defined by
\begin{equation}
    \bar m(x^i)=\sum_{o\in \mathcal{O}}\frac{e^{\alpha_o}}{\sum_{o'}e^{\alpha_{o'}}}o(x^i),
\end{equation}
where $o(x^i)$ denotes the output of operation $o$ applied to feature map $x^i$.
DARTS and DARTS-PT both attempt to solve the following expression via alternating gradient updates:

\begin{equation} \label{eq:darts}
\min_{\alpha}
\Lval \left(
\argmin_\theta \Ltrain \left( \theta, \alpha, \U \right), 
\alpha, \V \right).
\end{equation}

In particular, the gradient with respect to 
$\alpha$ can be approximated via
\begin{equation}
\nabla_{\alpha} \Lval \left(
\theta - \zeta\nabla_\theta\Ltrain \left( \theta, \alpha, \U \right), 
\alpha, \V \right),
\end{equation}
which can then be optimized using alternating gradient descent updates, according to a hyperparameter $\zeta$.

Once the supernetwork finishes training via gradient descent, the continuous NAS-parameters $\alpha$ must be \emph{discretized}.
In the original DARTS algorithm, this is achieved by taking the largest $\alpha_o$ on each edge.
However, \citet{wang2021rethinking} showed that this approach may not perform well. Instead, at each edge, DARTS-PT directly evaluates the  strength of each operation by its contribution to the supernetwork's performance, using a perturbation-based scoring technique~\citep{wang2021rethinking}.

\paragraph{Grad-Match}
Grad-Match, Gradient Matching based Data Subset Selection for Efficient
Deep learning Model Training is proposed in \cite{killamsetty2021grad}. Grad-Match selects a subset that best approximates either the full training dataset (or) a held-out validation dataset. This is achieved by selecting a coreset whose gradient matches the average loss gradient over the training dataset or the validation dataset respectively. The objective is modelled as a discrete subset selection problem that is combinatorially hard to solve and in response, they propose Orthogonal Matching Pursuit based greedy algorithm to pick up the subset.

The objective function for the Grad-Match version that selects a coreset to approximate training gradient is:
\begin{align}
    \underset{{S \subseteq \mathcal{U} |S| \leq k}}{\operatorname{argmin\hspace{0.7mm}}}
    \min_\mathbf{w} \mbox{E}_\lambda(\mathbf{w}, S) \text{ where, } \label{eq:gmobj}
\end{align}
where $\mathbf{w} \in \mathbb{R}^k$ represents the weight coefficient attached to each point in the coreset. Essentially, the formulation selects a subset whose weighted sum of gradients match the average training gradient.

\begin{align}
     \mbox{E}_\lambda & (\mathbf{w}, S) = \lambda \lVert \mathbf{w} \rVert^2  \\ \nonumber 
     & + \lVert {\sum_{j \in S} \mathbf{w}_j\nabla_\theta }\Ltrain \left( \theta, j \right) - \frac{1}{|\mathcal{U}|} {\sum_{u \in \mathcal{U}} \nabla_\theta }\Ltrain \left( \theta, u \right) \Vert
\end{align}

\paragraph{GLISTER.}
GLISTER, a Generalization based data subset selection for an efficient and robust learning framework, is a subset selection algorithm that selects a subset
of the training data, which maximizes the log-likelihood on a held-out validation set. This problem is formulated as a mixed discrete-continuous bi-level optimization problem.
GLISTER approximately solves the following
expression by first approximating the bi-level optimization expression using a single gradient step, 
and then using a greedy data subset selection procedure~\citep{killamsetty2020glister}.

\begin{equation} \label{eq:glister}
\min_{S\subseteq\U, |S|\leq k}
\Lval \left(
\argmin_\theta \Ltrain \left( \theta, S \right), 
 \V \right).
\end{equation}

In particular, the validation loss is approximated as follows:
\begin{align}
&\Lval \left(
\argmin_\theta \Ltrain \left( \theta, S \right), 
 \V \right)\\
\approx &\Lval \left(
\theta - \zeta\nabla_\theta\Ltrain \left( \theta, S \right), 
 \V \right). \label{eq:g_expression}
\end{align}

Thereafter, a simple greedy dataset subset selection procedure is employed to find the subset $S$ which approximately minimizes the validation loss~\citep{killamsetty2020glister}.

\paragraph{\model.}
There exist several possibilities for applying adaptive subset selection to one-shot NAS. We have considered two such possibilities (GLISTER and Grad-Match) and next, we present a formulation that organically combines Expressions~\eqref{eq:darts} and \eqref{eq:glister} into a single 
mixed discrete and continuous bi-level optimization problem.
The inner optimization is the minimization (over model weights $\theta$) of training loss during architecture training, on a subset of the training data of size $k$.
In the outer optimization, we minimize the validation loss by simultaneously optimizing over the 
NAS-parameters $\alpha$ as well as over the subset of the training data $S$. This optimization problem is aimed at efficiently determining the best (or at least an effective) neural architecture:




\begin{equation}
\argmin_{S\subseteq\U,|S|\leq k,\alpha}
\Lval \left(
\argmin_\theta \Ltrain \left( \theta, \alpha, S \right), 
\alpha, \V \right).
\end{equation}

Evaluating this expression is computationally prohibitive because of the expensive inner optimization problem.
Instead, we iteratively perform a joint optimization of the  weights $\theta$ from the inner optimization as well as the training subset $S$ and NAS-parameters $\alpha$ from the outer optimization. In order to iteratively update the training subset and architecture, we compute meta-approximations of the inner optimization.
As for the architecture, we compute
\begin{align}
&\nabla_{\alpha} \Lval \left(
\argmin_\theta \Ltrain \left( \theta, \alpha, S \right), 
\alpha, \V \right)\\
\approx &\nabla_{\alpha} \Lval \left(
\theta - \zeta\nabla_\theta\Ltrain \left( \theta, \alpha, S \right), 
\alpha, \V \right).
\end{align}
For the subset selection, following~\citet{killamsetty2020glister}, we run a greedy algorithm on a similar approximation to the inner optimization:
\begin{align}
&\Lval \left(
\argmin_\theta \Ltrain \left( \theta, \alpha, S \right), 
\alpha, \V \right)\\
\approx &\Lval \left(
\theta - \zeta\nabla_\theta\Ltrain \left( \theta, \alpha, S \right), 
\alpha, \V \right).
\end{align}

Then we can iteratively update the outer parameters (architecture and subset), and the inner parameters (weights).
Following prior work~\citep{killamsetty2020glister, darts}, 
we only update the architecture and subset every $t_1$ and $t_2$ steps, respectively, for efficiency ($t_1 << t_2$).
See Algorithm \ref{alg:glister-nas}.

We also tried GradMatch \citep{killamsetty2020glister}, an adaptive subset selection algorithm which finds subsets
that closely match the gradient of the training
or validation set, as our subset selection algorithnm and combined with DARTS-PT.

\begin{algorithm}[t]
    \centering
    \caption{\model}\label{alg:glister-nas}
    \begin{algorithmic}[1]
        \STATE \textbf{Require:} Training data $\U$, Validation data $\V$, Initial subset $S^{(0)}$ of size $k$, Initial parameters $\theta^{(0)}$ and $\alpha^{(0)}$, steps $T_1$, $T_2$, and $T$.
        \STATE \texttt{for all steps $t$ in $T$ do}
        \STATE \quad \texttt{if $t$ mod $T_1==0$:}
        \STATE \quad \quad $S^{(t)}=$ GreedyDSS$(\U,\V,\theta^{(t-1)},\alpha^{(t-1)})$
        \STATE \quad \texttt{else:}
        \STATE \quad \quad $S^{(t)} = S^{(t-1)}$
        \STATE \quad \texttt{if $t$ mod $T_2==0$:}
        \STATE \quad \quad Perform one step of SGD to update $\alpha^{(t)}$ using $V$
        \STATE \quad \texttt{else:}
        \STATE \quad \quad $\alpha^{(t)}=\alpha^{(t-1)}$
        \STATE \quad Perform one step of SGD to update $\theta^{(t)}$ using $S^{(t)}$ and $\alpha^{(t)}.$
        \STATE Discretize the supernet, 
        based on NAS-parameters $\alpha^{(T)}$ obtained using $S^{(T)}$, 
        to return final architecture 
        \STATE Train $\alpha$ using SGD with the full training set $\U$
        \STATE \textbf{Return:} Final architecture (discretized $\alpha^{(T)}$)
    \end{algorithmic}
\end{algorithm}

\paragraph{\adehb.}
Differential evolution hyperband (DEHB) \citep{awad2021dehb} is a leading algorithm for mutli-fidelity optimization which has been applied to both hyperparameter optimization (HPO) and NAS \citep{awad2021dehb, vincent2022improved}.
The approach combines differential evolution \citep{storn_de}, a population-based evolutionary algorithm, with hyperband \citep{hyperband}, a bandit-based multi-fidelity optimization routine which rules out poor hyperparameter settings before they are trained for too long.
Unlike DARTS-based approaches, DEHB does not use a supernetwork -- each architecture is trained separately. 
Therefore, to devise \adehb, we incorporate adaptive subset selection simply by running GLISTER for each individual architecture trained throughout the algorithm.

\paragraph{\abohb.}
Bayesian Optimization and Hyperband (BOHB), \citep{falkner2018bohb} is a hyperparameter optimization method  that combines benefits of Bayesian Optimization and bandit based methods \citep{hyperband} such that it finds good solutions  faster than Bayesian optimization and converges to the best solutions  faster than Hyperband. We use adapative subset selection along with BOHB to devise \abohb{}
which gives almost similar accuracy of BOHB while reducing the runtime significantly.

%% file: experiments.tex
\section{Experiments} \label{sec:experiments}


In this section, we describe our experimental setup and results.

\paragraph{Search spaces.}
We perform experiments on NAS-Bench-201 with CIFAR-10, CIFAR-100, and ImageNet16-120,
DARTS with CIFAR-10, and DARTS-S4 with CIFAR-10. 

NAS-Bench-201 \citep{nasbench201} is a cell-based search space which contains 15\,625 architectures, or 6\,466 architectures that are unique up to isomorphisms.
Each cell is a directed acyclic graph consisting of four nodes and six edges. Each of the six edges have five choices of operations. 
The cell is then stacked several times to form the final architecture.

The DARTS search space~\citep{darts} is a cell-based search space containing $10^{18}$ architectures. It consists of a normal cell and a reduction cell, each of which is represented as a directed acyclic graph with four nodes and two incoming edges per node. Each edge has eight choices of operations, and a choice of input node.
Similar to NAS-Bench-201, the cells are stacked several times to form the final architecture.

\citet{zela2020understanding} propose a variant S4 of the DARTS search space, which replaces the original set of eight choices of operations with just two operations, \viz: $3\times 3$ \emph{SepConv}, and \emph{Noise}, where \emph{Noise} replaces the feature map values by noise drawn from $\epsilon \sim\mathcal{N}(0,1)$.
This search space was designed to test the failure modes of one-shot NAS methods such as DARTS; it is expected that \emph{Noise} is not chosen, since it would actively hurt performance. S4 and DARTS have no differences other than the operation set. 

\paragraph{Methods tested.}
We perform experiments with DARTS-PT, \model, and three other (non-adaptive) data subset selection methods applied to DARTS-PT.
We describe the details of each approach below. 
\begin{itemize}[topsep=0pt, itemsep=2pt, parsep=0pt, leftmargin=5mm]
    \item \dpt{}: We use the original implementation of DARTS-PT~\citep{wang2021rethinking} as described in  Section~\ref{sec:methodology}. 
    \item \dptrand{}: This is similar to \dpt{}, but the supernetwork is trained and discretized using 
    a random subset of the training data.
    \item \dptfl{}: While similar to \dpt{}, the supernetwork is trained and discretized using 
    a subset of the training data, selected using facility location. Facility location function tries to find a representative subset of items. The Facility-Location function is similar to k-medoid clustering. For each data point $i$ in the ground set $V$, we compute the representative from subset $X$ which is closest to $i$ and add these similarities for all data points. Facility-Location is monotone submodular.
    \begin{equation}
        f(X) = \sum_{i\in\mathcal{V}}\max_{j \in \mathcal{X}}s_{ij}
    \end{equation}
    
    The facility location algorithm was implemented using the naive greedy algorithm and run on each class separately, using a dense Euclidean metric.
    For this, we employed the \texttt{submodlib} library~\citep{kaushal2022submodlib}.
    \item \dptent{}~\citep{na2021accelerating}: Again this bears similarity to \dpt{} but with a difference. The supernetwork is trained and discretized using a subset of the training data, 
    selected using a combination of high and low-entropy datapoints.
    The cost of NAS is reduced by selecting a representative set of the original training data. Unlike the other existing zero cost subset selection methods for NAS, this approach is specifically tailored for NAS and accelerates neural architecture search using proxy data. The entropy of a datapoint is calcuated by training a base neural architecture from the searce space, and determining whether the output probability is low or high.
    This approach was adopted by~\citet{na2021accelerating} to speed up DARTS.
    \item \model: This is our approach, as described in the previous section; more specifically, see Algorithm \ref{alg:glister-nas}.
\end{itemize}

\paragraph{Experimental setup.}
Following~\citet{wang2021rethinking}, we use 50\% of the full training dataset for supernet training and 50\% for validation. We report the accuracy of the finally obtained architecture on the held-out test set.
In our primary  experiments, for each (adaptive or non-adaptive) subset selection method, we set the subset size to 10\% of the training dataset.
We run the same experimental procedure for each method: select a size-10\% subset of the full training dataset, train and discretize the supernet on the subset, and train the final architecture using the full training dataset. 
For \dpt{}, we run the same procedure using the full training dataset at each step.
We otherwise use the exact same training pipeline as in~\citet{wang2021rethinking}, \viz, batch size of 64, learning rate of 0.025, momentum of 0.9, and cosine annealing.

We run all experiments on an NVIDIA Tesla V100 GPU.
We run each algorithm with 5 random seeds, reporting the mean and standard deviation of each method, with the exception of \dpt{}; due to its extreme runtime and
availability of existing results, we perform the experiment once and verify that the result is nearly identical to published results~\citep{wang2021rethinking}.
We also report the time it takes to output the final architecture.

\paragraph{Experimental results and discussion.}
In Tables \ref{tab:cifar10}, \ref{tab:cifar100}, and \ref{tab:imagenet}, 
we report the results on NAS-Bench-201.
On CIFAR-10 and ImageNet16-120, \model{} yields significantly higher accuracy than all other algorithms tested. On CIFAR-100, \model{}  is essentially tied with \dptfl{} for the highest accuracy.
Furthermore, all NAS algorithms that use subset selection have significantly decreased runtime -- \model{} sees a factor of 9 speedup compared to \dpt{}.
Note that \dptfl{} takes more time when the number of examples per class in the dataset is higher, so it sees comparatively higher runtimes on CIFAR-10.

In Tables \ref{tab:s4} and \ref{tab:darts}, we report the results on S4 CIFAR-10 and DARTS CIFAR-10.
Once again, the runtime of \model{} is significantly faster than \dpt{} -- a factor of 7 speedup.
On these search spaces, the performances of the subset-based methods are more similar when compared to NAS-Bench-201, and on the DARTS search space, \model{} does not outperform \dpt{}.
A possible explanation is that S4 and DARTS are significantly larger search spaces than NAS-Bench-201 and require more training data to distinguish between architectures.
To test this, we included an additional  experiment in Table \ref{tab:darts}, giving \model{} 20\% training data instead of 10\%. We find that the accuracy significantly increases, moving within one standard deviation of the accuracy of \dpt{}.

In Table 6, we report the results of DARTS-PT with adaptive subset selection using Grad-Match \citep{killamsetty2021grad} on NAS-Bench-201 with datasets CIFAR-10 and CIFAR-100. Although DARTS-PT with Grad-Match was not able to beat the scores of \model{}, it still gave better results than most non-adaptive subset selection methods.

Overall, \model{} achieves the highest average performance across all search spaces.
Furthermore, \model{} achieves no less than a seven-fold increase in runtime compared to \dpt{}, on all search spaces.

We also tried the combination of DEHB~\citep{awad2021dehb} with Adaptive Subset selection (GLISTER).  A configuration sampled from a parameter space (with parameters such as Kernel size, channel size, stride) is used to instantiate a CNN architecture (we used the same architecture as in ~\citep{awad2021dehb}). On this architecture, we trained DEHB on the MNIST dataset for 100 epochs with and without subset selection. When tested on five different seeds, DEHB trained without adaptive subset selection took   \textbf{0.91} hours and gave \textbf{0.96 $\pm$ 0.03} accuracy whereas \adehb{} using $20\%$ data and selecting subset at every 10 epochs took \textbf{0.64} hours and yielded \textbf{0.99 $\pm$ 0.00} accuracy.

We used BOHB for MNIST dataset and ran for 100 epochs. We used 32k training and 8k validation datapoints. One set of experiments was done with this complete data and another with a subset of these selected by GLISTER every 10 epochs.
BOHB without adaptive subset selection gave an accuracy of \textbf{0.99 $\pm$ 0.00} and took \textbf{2.43} hours on MNIST dataset whereas \abohb{} gave an accuracy of \textbf{0.98 $\pm$ 0.00} and took \textbf{1.16} hours with $20\%$ data and selecting subset at every 10 epochs.

\begin{figure}
     \centering
     \begin{subfigure}[b]{0.22\textwidth}
         \centering
         \includegraphics[width=\textwidth]{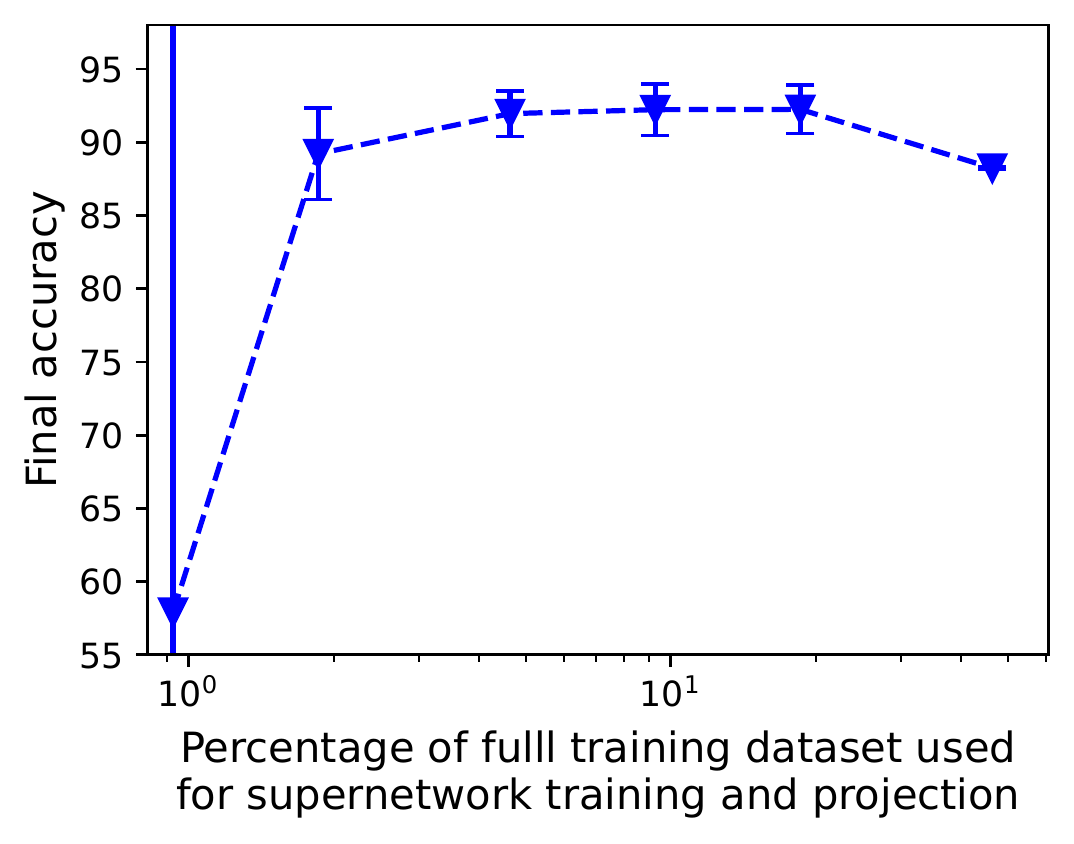}
     \end{subfigure}
     \hfill
     \begin{subfigure}[b]{0.22\textwidth}
         \centering
         \includegraphics[width=\textwidth]{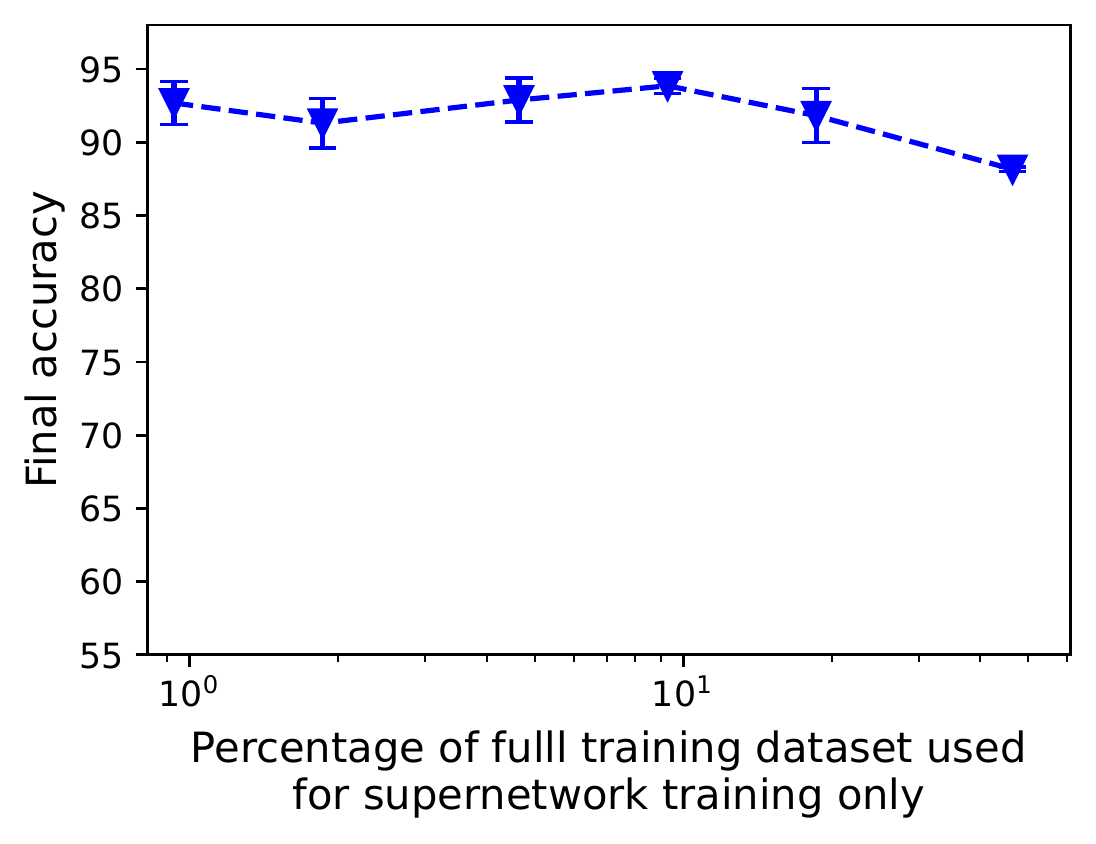}
     \end{subfigure}
     \centering
     \begin{subfigure}[b]{0.22\textwidth}
         \centering
         \includegraphics[width=\textwidth]{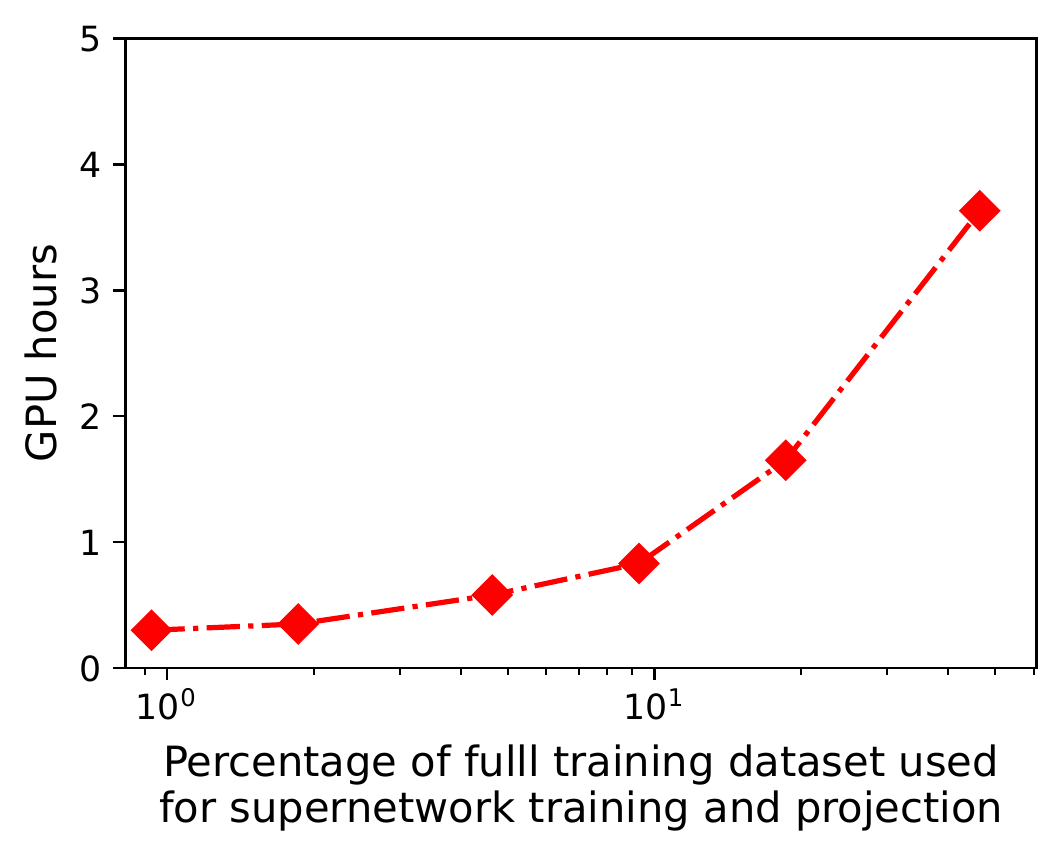}
     \end{subfigure}
     \hfill
     \begin{subfigure}[b]{0.22\textwidth}
         \centering
         \includegraphics[width=\textwidth]{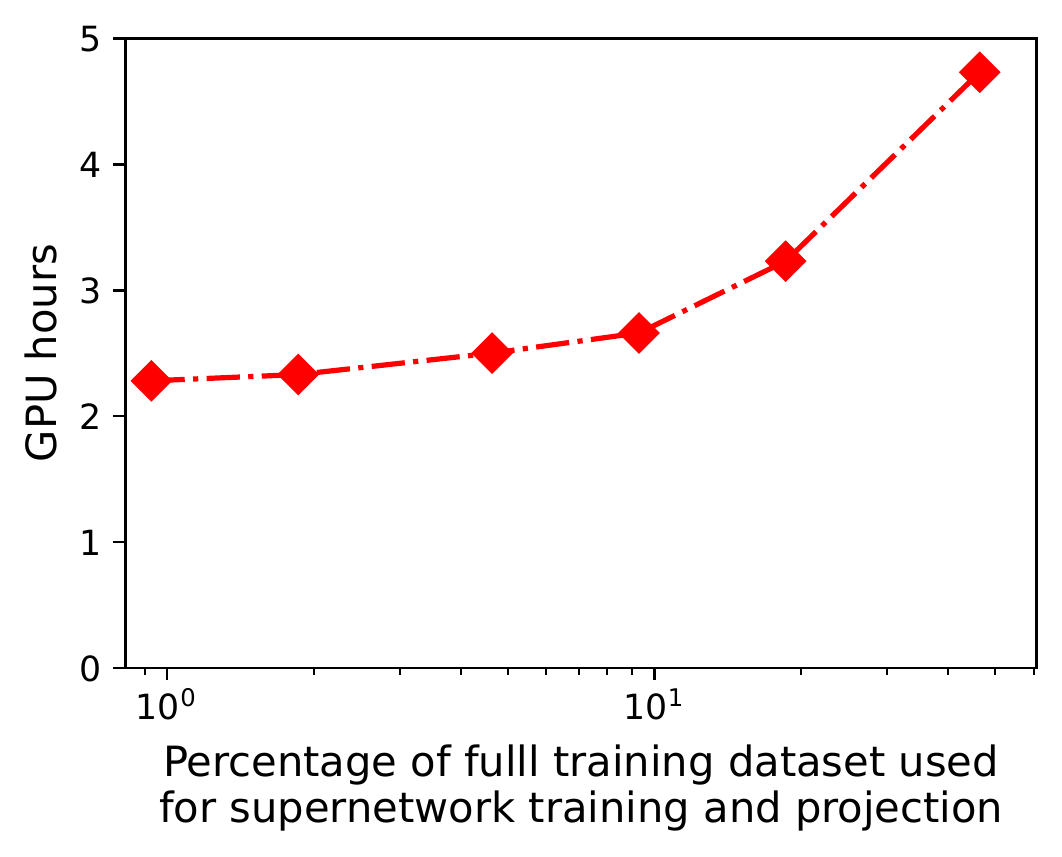}
     \end{subfigure}
        \caption{Performance and runtime of \model{} varies as the percentage of training data increases.
        (Left) The supernetwork training and projection step are given a percentage of the full training dataset. 
        (Right) The supernetwork training is given a percentage, while the projection step is given the full training dataset.}
        \label{fig:ablation}
\end{figure}

\paragraph{Ablation study.}
To explore the effect of the percentage of data used, in Figure \ref{fig:ablation} (left), we run \model{} with different percentages of the training data, ranging from 1\% to 50\%.
In the Figure~\ref{fig:ablation} (right), we run the same experiment using the full training data in the projection step of \model.
Interestingly, we see a definitive U-shape in the first experiment: the highest accuracy with \model{} is at 20\%, achieving accuracy \emph{higher} than the standard setting of 100\% data (\ie, \dpt{}). Since the supernetwork is an over-parameterized model of weights and architecture parameters, and \model{} regularly updates the training subset to maximize validation accuracy, \model{} may help prevent the supernetwork from overfitting. 
Furthermore, in the second experiment, we see that  relatively, the accuracies are much more consistent when varying the percentage of the training set used, when the projection step is allowed to use the full training dataset. Therefore, keeping the full training dataset for the projection step leads to higher and more consistent performance, at the expense of more GPU-hours. 

Overall, based on the ablation studies in Figure~\ref{fig:ablation}, the user may decide their desired tradeoff between performance and accuracy, and choose the subset size in the supernetwork training accordingly. For example, with a budget of 1 GPU hour, the best approach is to use a 10\% subset of the training data for the supernet training and projection, but with a budget of 2.5 GPU hours, the best approach is to use a 10\% subset of the training data for the supernet and the full training data for the projection.

\begin{table*}[]
    \caption{Performance of one-shot NAS algorithms on NAS-Bench-201 CIFAR-10.}
    \label{tab:cifar10}
    \centering
\begin{tabular}{ |p{3.5cm}||p{2.5cm}|p{2cm}|p{2cm}|  }
 \hline
 \multicolumn{4}{|c|}{Performance on NAS-Bench-201 CIFAR-10} \\
 \hline
 Algorithm & Test accuracy & GPU hours & \% Data used\\
 \hline
 \dpt{}   &  88.21 (88.11)   & 7.50 &  100\\
 \hline
 \dptent{}   &  $86.31\pm 4.66$  & \textbf{0.62} & 10\\
 \hline
 \dptrand{}   &  $86.94\pm 3.58$  & \textbf{0.62} & 10\\
 \hline
 \dptfl{} & $89.27\pm 1.09$ & 1.60 & 10\\
 \hline
 \model{} & $\textbf{92.22}\pm 1.76$ & 0.83 & 10\\
 \hline
\end{tabular}
\end{table*}

\begin{table*}[]
    \caption{Performance of one-shot NAS algorithms on NAS-Bench-201 CIFAR-100.}
    \label{tab:cifar100}
    \centering
\begin{tabular}{ |p{3.5cm}||p{2.5cm}|p{2cm}|p{2cm}|  }
 \hline
 \multicolumn{4}{|c|}{Performance on NAS-Bench-201 CIFAR-100} \\
 \hline
 Algorithm & Test accuracy & GPU hours & \%Data used \\
 \hline
 \dpt{}   &  61.650   & 8.00 &   100\\
 \hline
 \dptent{}   &   $56.79\pm 7.63$   & \textbf{0.58} & 10\\
 \hline
 \dptrand{}   &  $56.95\pm 4.54$   & \textbf{0.58} & 10 \\
 \hline
 \dptfl{} & $\textbf{64.28}\pm 3.10$  & 0.67 & 10\\
 \hline
 \model{} & $64.27\pm 3.37$   & 0.87 & 10\\
 \hline
\end{tabular}
\end{table*}

\begin{table*}[]
    \caption{Performance of one-shot NAS algorithms on NAS-Bench-201 ImageNet16-120.}
    \label{tab:imagenet}
    \centering
\begin{tabular}{ |p{3.5cm}||p{2.5cm}|p{2cm}|p{2cm}|  }
 \hline
 \multicolumn{4}{|c|}{Performance on NAS-Bench-201 Imagenet16-120} \\
 \hline
 Algorithm & Test accuracy & GPU hours & \%Data used\\
 \hline
 \dpt{}   &  35.00   & 33.50  & 100\\
 \hline
 \dptent{}   &  $26.52\pm 3.73$   & \textbf{1.58} & 10\\
 \hline
 \dptrand{}   &  $27.04\pm 5.53$   & \textbf{1.58}  & 10 \\
 \hline
 \dptfl{} & $29.30\pm 5.35$ & 1.90 & 10\\
 \hline
 \model{} & $\textbf{36.10}\pm 6.96$    & 2.60  & 10\\
 \hline
\end{tabular}
\end{table*}

\begin{table*}[]
    \caption{Performance of one-shot NAS algorithms on S4 search space CIFAR-10.}
    \label{tab:s4}
    \centering
\begin{tabular}{ |p{3.5cm}||p{2.5cm}|p{2cm}|p{2cm}|  }
 \hline
 \multicolumn{4}{|c|}{Performance on S4 CIFAR-10} \\
 \hline
 Algorithm & Test accuracy & GPU hours & \%Data used\\
 \hline
 \dpt{}   &  97.31 (97.36)  & 8.38  & 100\\
 \hline
 \dptent{}   &  $\textbf{97.45}\pm 0.10$   & \textbf{0.86} & 10\\
 \hline
 \dptrand{}   &  $97.40\pm 0.06$   & \textbf{0.86}  & 10 \\
 \hline
 \dptfl{} & $97.34\pm 0.13$ & 1.08 & 10\\
 \hline
 \model{} & $97.30\pm 0.12$    & 1.08  & 10\\
 \hline
\end{tabular}
\end{table*}

\begin{table*}[]
    \caption{Performance of one-shot NAS algorithms on DARTS search space CIFAR-10.}
    \label{tab:darts}
    \centering
\begin{tabular}{ |p{3.5cm}||p{2.5cm}|p{2cm}|p{2cm}|  }
 \hline
 \multicolumn{4}{|c|}{Performance on DARTS CIFAR-10} \\
 \hline
 Algorithm & Test accuracy & GPU hours & \%Data used\\
 \hline
 \dpt{}  &  \textbf{97.17} (97.39)  & 20.59  & 100\\
 \hline
 \dptent{}   &  $96.68\pm 0.26$   & 3.40 & 10\\
 \hline
 \dptrand{}   &  $97.01\pm 0.32$   & \textbf{2.35}  & 10 \\
 \hline
 \dptfl{} & $96.91\pm 0.15$ & 4.00 & 10\\
 \hline
 \model{} & $96.73\pm 0.29$    & 2.75  & 10\\
 \hline
 \model{} & $96.97\pm 0.24$    & 4.50  & 20\\
 \hline
\end{tabular}
\end{table*}

\begin{table*}[]
    \caption{Performance of DARTS-PT + GRAD-MATCH on NAS-Bench-201 }
    \label{tab:cifar10-grad}
    \centering
\begin{tabular}{ |p{3.5cm}||p{3cm}|p{3cm}|p{2.5cm}|  }
 \hline
 \multicolumn{4}{|c|}{Performance on NAS-Bench-201 CIFAR-10} \\
 \hline
 Dataset & Test accuracy & GPU hours & \% Data used\\
 \hline
 CIFAR-10   & $88.83\pm 1.09$   & 0.87 &  10\\
 \hline
 CIFAR-100   & $63.70\pm 3.98$   & 0.87 &  10\\
 \hline

\end{tabular}
\end{table*}


%% file: conclusion.tex
\section{Conclusions, Limitations, and Impact} \label{sec:conclusion}

In this work, we used a connection between one-shot NAS algorithms and adaptive subset
selection to devise an algorithm that makes use of state-of-the-art techniques from both areas. Specifically, we build on DARTS-PT and GLISTER, that are state-of-the-art approaches to one-shot NAS and adaptive subset selection, respectively, and pose a bi-level optimization problem on the training and validation sets. This leads us to the formulation of  a combined approach, \viz, \model, as a mixed discrete and continuous bi-level optimization problem.
We empirically demonstrated that the resulting algorithm is able to train on an (adaptive) dataset that is 10\% of the size of the full training set, without sacrificing accuracy, resulting in an order of magnitude decrease in runtime. We also show how this method can be extended to hyperparameter optimization algorithms, in general, using \adehb{} and \abohb{}.
We also release a codebase consisting of four different subset selection techniques integrated into one-shot NAS and profiled on the different benchmarks.\

\paragraph{Limitations.}
While \model{} uses a subset of the data when training and discretizing 
the supernetwork, the full dataset is used for training the final architecture.
Another interesting direction for future work is to use an adaptive subset of the data even
when training the final architecture, which may lead to even faster runtime, 
perhaps at a small cost to performance.

Another interesting direction for future work is to apply adaptive subset selection 
to other non supernet-based NAS algorithms such as regularized evolution \citep{real2019regularized} or BANANAS \citep{bananas}.
Since GLISTER is able to significantly reduce the runtime to train architectures, it would be expected that GLISTER can be used to reduce the runtime of regularized evolution, BANANAS, and other iterative optimization-based NAS algorithms by up to an order of magnitude.

\paragraph{Broader impact.}
Our work combines techniques from two different areas: adaptive subset selection for machine learning, 
and neural architecture search. The goal of our work is to make it easier and quicker to develop 
high-performing architectures on new datasets.
Our work also helps to unify two sub-fields of machine learning that had thus far been disjoint.
There may be even more opportunity to use tools from one sub-field to make progress in the other
sub-field, and our work is the first step at bridging these subfields.

Since the end product of our work is a NAS algorithm, it is not itself meant for one application
but can be used in any end-application.
For example, it may be used to more efficiently find deep learning architectures for applications that 
help to reduce CO2 emissions, or for creating large language models.
Our hope is that future AI models discovered by our work will have a net positive impact,
due to the push for the AI community to be more conscious about the societal impact of its
work \citep{hecht2018time}.

%% file: appendix.tex


\section{Additional Results and Analyses}
In this section, we give additional results and analyses to supplement Section \ref{sec:experiments}.

In Table \ref{tab:summary}, we give a summary of the improvements of \model{}  when compared to \dpt{}.
\begin{table*}[]
    
    \caption{Summary of Improvements over \dpt{} by \model{} }
    \label{tab:summary}
    \centering
\begin{tabular}{||p{2.5cm}||p{2.5cm}|p{2.5cm}|p{2cm}|p{2cm}||} 
 \hline
 Search Space & Dataset & Accuracy & Time reduced &  \% Data Used \\ [0.4ex] 
 \hline
 NAS-Bench-201 & CIFAR-10 & +5.07 & 8.62 times & 10 \\ 
\hline
 NAS-Bench-201 & CIFAR-100 & +2.63 & 9.20 times & 10 \\ 
\hline
 NAS-Bench-201 & Imagenet-16-120 & +1.10 & 12.80 times & 10 \\ 
 \hline
 S4 & CIFAR-10 & -0.01 & 7.76 times & 10 \\ 
  \hline
 DARTS & CIFAR-10 & -0.44 & 7.49 times & 10 \\ 
 \hline
 DARTS & CIFAR-10 & -0.20 & 4.58 times & 20 \\ 
 \hline
\end{tabular}
\end{table*}


In Table \ref{tab:cifar10-app}, we give the results of \model{} with using the full data for the DARTS-PT projection step for search spaces DARTS and S4. Although we were able to get better accuracy (when compared 10\% data on projection step) on DARTS space, the accuracy went down a little bit for S4.

\begin{table}[]
    
    \caption{Performance of \model{} on other search spaces}
    \label{tab:cifar10-app}
    \centering
\begin{tabular}{ |p{1.8cm}||p{1.8cm}|p{1.5cm}|p{1.8cm}|  }
 \hline
 \multicolumn{4}{|c|}{Performance on CIFAR-10} \\
 \hline
 Search Space & Test accuracy & GPU hours & \% Data used\\
 \hline
 DARTS   & $97.06\pm 1.09$   & 2.65 &  10\\
 \hline
 S4   & $97.28\pm 0.03$   & 1.18 &  10\\
 \hline

\end{tabular}
\end{table}


In Tables \ref{tab:subset-ablation} and \ref{tab:projection-ablation}, we give the tables to match the plots in Figure \ref{fig:ablation}.

\begin{table}[]
    \caption{Ablation study of \model{} on NAS-Bench-201 search space CIFAR-10.}
    \label{tab:subset-ablation}
    \centering
\begin{tabular}{ |p{2.4cm}|p{2.4cm}|p{2.5cm}|  }
 \hline
 \multicolumn{3}{|c|}{Performance on NAS-Bench-201 CIFAR-10} \\
 \hline
 Test accuracy & GPU hours & \%Data used\\
 \hline
 $57.91\pm 43.77$    & 0.30  & 1\\
 \hline
 $89.21\pm 3.12$    & 0.35  & 2\\
 \hline
 $91.95\pm 1.56$    & 0.58  & 5\\
 \hline
 $92.22\pm 1.76$    & 0.83  & 10\\
 \hline
 $\textbf{92.24}\pm 1.65$    & 1.58  & 20\\
 \hline
 $88.23\pm 0.06$    & 3.63  & 50\\
 \hline
\end{tabular}
\end{table}

\begin{table}[]
    \caption{Ablation study of \model{} on NAS-Bench-201 search space CIFAR-10 with the 
    full data for the \dpt{} projection.}
    \label{tab:projection-ablation}
    \centering
\begin{tabular}{ |p{2.4cm}|p{2.4cm}|p{2.5cm}|  }
 \hline
 \multicolumn{3}{|c|}{Performance on NAS-Bench-201 CIFAR-10} \\
 \hline
 Test accuracy & GPU hours & \%Data used\\
 \hline
 $92.68\pm 1.47$    & 2.28  & 1\\
 \hline
 $91.30\pm 1.69$    & 2.33  & 2\\
 \hline
 $92.88\pm 1.51$    & 2.50  & 5\\
 \hline
 $\textbf{93.85}\pm 0.51$    & 2.66  & 10\\
 \hline
 $91.81\pm 1.84$    & 3.23  & 20\\
 \hline
 $88.15\pm 0.14$    & 4.73  & 50\\
 \hline
\end{tabular}
\end{table}

In Table \ref{tab:cifar100-app}, we give the results of \model{} using the full data for the \dpt{} projection (perturbation) step for search space NAS-Bench-201 for datasets CIFAR-100 and Imagenet16-120. Although we were able to get better accuracy (when compared to \model{} without full data for the perturbation step), the time taken was significantly more.

\begin{table}[]
    
    \caption{Performance of \model{} on NAS-Bench-201}
    \label{tab:cifar100-app}
    \centering
\begin{tabular}{ |p{1.8cm}||p{1.8cm}|p{1.5cm}|p{1.8cm}|  }
 \hline
 \multicolumn{4}{|c|}{Performance on CIFAR-10} \\
 \hline
 Dataset & Test accuracy & GPU hours & \% Data used\\
 \hline
 CIFAR-100   & $65.85\pm 4.17$   & 2.43 &  10\\
 \hline
 Imagenet16-120   & $37.43\pm 2.12$   & 8.78 &  10\\
 \hline

\end{tabular}
\end{table}

\subsection{Results of DARTS-PT-GRAD-MATCH}

In Section \ref{sec:methodology}, we introduced GLISTER applied to DARTS-PT as \model.
However, there is another choice of adaptive subset selection algorithm: GRAD-MATCH \citep{killamsetty2021grad}.
In this section, we describe GRAD-MATCH and give results on GRAD-MATCH applied to DARTS-PT, showing that it does not work as well as \model.

The core idea of GRAD-MATCH is to find a subset of the original training set whose gradients match the gradients of the training/validation set. The gradient error term can be given as
\begin{equation}
Err(w^t, X^t, L, L_T, \theta_t) = \left\lVert \sum_{i \in X^t} w_i^t\nabla_\theta L^i_T(\theta_t) - \nabla_\theta L(\theta_t)\right\rVert
\end{equation}

where $w^t$ are the weights produced by the adaptive data selection algorithm,
$X^t$ are the subsets selected,
$L_t$ is the training loss,
$L$ is either the training or the validation loss, and
$\theta_t$ are the classifier model parameters.
In GRAD-MATCH, minimizing the above equation is reformulated as optimizing an equivalent submodular function with approximations for efficiency.

GRAD-MATCH was integrated with \dpt{} the same way we described integrating GLISTER with \dpt{} in Section \ref{sec:methodology}. Furthermore, we used the same hyperparameters for \dpt{} as with \model. 
We denote the algorithm as DARTS-PT-GRAD-MATCH.
The results were found to be better than some baselines but still not better than \model{} for CIFAR-10 and CIFAR-100 on NAS-Bench-201. While GRAD-MATCH is faster compared to GLISTER, the bottleneck in one-shot NAS is training the supernetwork, therefore, the improved performance of GLISTER makes it the better fit to be incorporated with \dpt{} in creating \model.


\section{Details from Section \ref{sec:experiments}}

In this section, we give more details for the experiments conducted in Section \ref{sec:experiments}.

\subsection{Experiments on NAS-Bench-201}
We used the original code from \dpt{} \citep{wang2021rethinking} and GLISTER \citep{killamsetty2020glister}. The \dpt{} code consists of two parts. The supernet training and a perturbation based projection part to discretize $\alpha$. The Supernet training is run for 100 epochs and at each 10 epoch interval, we select a new subset of data by passing the model and architecture parameters. At every epoch, we use 10\% of the original dataset. We use a batchsize of 64, learning rate of 0.025, momentum of 0.9, and cosine annealing. We use 50\% of data for training and 50\% for validation, as in the DARTS-PT paper \citep{wang2021rethinking}. The last 10\% data subset is saved and used for the perturbation based projection part of \dpt{}. We run the projection part for 25 epochs. For subset selection, we used the same code of GLISTER with selection algorithm run on each class separately.

For \dptfl{}, we used the implementation of Facility Location as present in \texttt{submodlib}.
This subset selection algorithm was used in the dense Euclidean setting.
The algorithm is used separately for each class so as to keep the representation across classes the same as original.
It was optimised using the `NaiveGreedy' algorithm,
For the experiments, 10\% data was used.

For \dptrand{} and \dptent{}, we combined \dpt{} with proxy data using two methods of subset selection techniques for dataset, one a random subset selection and other an entropy based subset selection technique \citep{na2021accelerating}. 
For random subset data was choosen randomly from the dataset. 
For the entropy based selection, we used the entropy files for CIFAR-10 and CIFAR-100 from Na et al.\ \citep{na2021accelerating} 
which was obtained by training a baseline network of ResNet20 and ResNet56 respectively. For ImageNet16-120, we trained a ResNet-50 model from the PyTorch model zoo. 

For S4 and the DARTS search space, we used the same configuration as for NAS-Bench-201. Since S4 and DARTS are non-tabular, we used a separate evaluation code for computing the performance of the selected architecture. We used the same evaluation code given in \dpt{}. The code uses a batch size of 96, learning rate of 0.025, momentum of 0.9 and weight decay of 0.025. The architecture is trained for 600 epochs. 



\section{Additional Details of the Search Spaces}
\subsection{NAS-Bench-201}
In NAS-Bench-201 \citep{nasbench201} the search space is based on cell-based architectures where each cell is a DAG. Here each node is a feature map and each edge is an operation. The search space for NAS-Bench-201 is defined by 4 nodes and 5 operations making 15625 different cell candidates.

NAS-Bench-201 gives performance of every candidate architecture on three different datasets (CIFAR-10, CIFAR-100, Imagenet-16-120).  This makes NAS-Bench-201 a fair benchmark for the comparison of different NAS algorithms. The five representative operations chosen for NAS-Bench-201 are: (1) zeroize (dropping the associated edge) (2)skip connection (3) 1-by-1 convolution (4) 3x3 convolution (5) 3x3 average pooling layer. Each convolution operation is a sequence of ReLU, convolution and batch normalization. The input of each node includes the sum of all the feature maps transformed using the respective edge operations. Each candidate architecture is trained using Nestorov momentum SGD using cross entropy loss for 200 epochs.

\subsection{DARTS-CNN search space}

The search space is represented using cell based architectures \citep{darts}. Each cell is a DAG with feature maps as nodes and edges as operations. The operations included are 3x3 and 5x5 separable convolutions, 3x3 and 5x5 dilated separable convolutions, 3x3 max pooling, 3x3 average pooling, identity and zero. Each cell consists of 7 nodes where output node is depth-wise concatenation of all the intermediate nodes.

\subsection{DARTS S4}

S1-S4 are four different search spaces proposed by Zela et al.\ \citep{zela2020understanding}. These search spaces were proposed to demonstrate the failure of standard DARTS. The same micro-architecture as the original DARTS paper with normal and reduction cells is used but only a subset of operators are allowed for the search spaces. The representative set of operations for S4 is \{3x3 SepConv, Noise\}. SepConv is chosen since it is one of the most common operation in the discovered cells reported by Liu et al.\ \citep{darts}. Noise operation plugs in the noise values $\epsilon \sim N(0, 1)$ for every value from the input feature map.

\section{Additional Details of the Algorithms Implemented}
In this section, we give more details for GLISTER and the baselines used in Section \ref{sec:experiments}.

\subsection{Details of GLISTER}
The optimization that we are trying to solve for GLISTER, equation \eqref{eq:glister}, can be written as
\begin{equation} \label{eq:13}
    S^{t+1} = argmin_{S \subseteq U, |S|\leq k} G_{\theta^t}(S)
\end{equation}

where $G_{\theta^t}(S)$ is equation \ref{eq:g_expression}. Since equation \eqref{eq:13} is an instance of submodular optimization (as proven in Theorem 1 of \citep{killamsetty2020glister}), it can be regularized using another function such as supervised facility location. The regularized objective can be written as
\begin{equation} \label{eq:14}
    S^{t+1} = \argmin_{S \subseteq U, |S|\leq k} G_{\theta^t}(S) + \lambda R(S)
\end{equation}
where $\lambda$ is a trade-off parameter.
GreedyDSS refers to the set of greedy algorithms and approximations that solves \ref{eq:14}.
Greedy Taylor Approximation algorithm (GreedyTaylorApprox(U, V, $\theta^0$, $\eta$, k, r, $\lambda$, R), described as Algorithm 2 in \citep{killamsetty2020glister}) is used as GreedyDSS in our work.
Here, $U$ and $V$ are the training and validation set respectively. $\theta^t$ is the current set of parameters, $\eta$ is the learning rate, $k$ is the budget, parameter $r$ governs the number of times we perform the Taylor series approximation, and $\lambda$ is the regularization constant.


\subsection{Details of facility location}
Intuitively, facility location, attempts to model representation of the datapoints. 
If $s_{ij}$ is the similarity between datapoints $i$ and $j$, define $f(X)$ such that

\begin{equation}
    f(X) = \underset{i \in V}{\sum} \underset{j \in X}{\max} s_{ij}
\end{equation}

where $V$ is the ground set. If the ground set of items are assumed clustered, an alternative clustered implementation of facility location is computed over the clusters as
\begin{equation}
    f(X) = \underset{l \in 1..K}{\sum}\underset{i \in C_l}{\sum} \underset{j \in X \cap C_l}{\max} s_{ij}
\end{equation}

\subsection{Details of DARTS-PT-ENTROPY}
\dptent{} is the implementation of \citep{na2021accelerating} where the cost of NAS is reduced by selecting a representative set of the original training data.
The entropy of a datapoint is defined as
\begin{equation}
\text{Entropy}(x: M) = -\sum_{\tilde{y}}p(\tilde{y}|x, M)\log p(\tilde{y}|x, M)
\end{equation}
where $\tilde{y} = M(x)$ is the predictive distribution of $x$ w.r.t.\ a pre-trained baseline model $M$.


The selection method selects datapoints from both the high entropy and low entropy regions.
%

If $h_x$ is a bin of the data point $x$ on data entropy histogram $H$, $|h_x|$ is the height of $h_x$ (number of examples in $h_x$), three probabilities are defined as
\begin{equation}
P_{\{1,2,3\}}(x;H) = \text{norm}(W_{\{1,2,3\}}(h_x;H)/|h_x|)
\end{equation}
where $\text{norm}()$ is a normalizer such that the probability terms add to 1.
$W_{\{1,2,3\}}$ are selected such that the tail end entropy data are likely to be selected over the middle entropy data points.

In \citep{na2021accelerating}, $P_1(x)$ was the highest performer. We have used $P_1(x)$ in our experiments.
